\documentclass{article}

\PassOptionsToPackage{numbers,sort&compress}{natbib}

\usepackage[preprint]{neurips_2026}

\usepackage[utf8]{inputenc} 
\usepackage[T1]{fontenc}    
\usepackage{hyperref}       
\usepackage{url}            
\usepackage{booktabs}       
\usepackage{amsfonts}       
\usepackage{nicefrac}       
\usepackage{microtype}      
\usepackage{xcolor}         
\usepackage{graphicx} 
\usepackage{enumitem}
\usepackage{amsmath}
\usepackage{wrapfig}
\usepackage{amssymb}
\usepackage{mathtools}
\usepackage{amsthm}
\usepackage{multirow}
\usepackage{algorithm}
\usepackage{algpseudocode}

\title{Test Time Training for Supervised Causal Learning}

\author{%
  Zizhen Deng\thanks{The work was done during his internship at Microsoft.} \\
  Peking University\\
  \texttt{zzdeng25@stu.pku.edu.cn} \\
  \And
  Jiaru Zhang \\
  Shanghai Jiao Tong University \\
  \texttt{jiaruzhang@sjtu.edu.cn} \\
  \AND
  Rui Ding\thanks{Corresponding author.} \\
  Microsoft \\
  \texttt{juding@microsoft.com} \\
  \And
  Huang Bojun \\
  Sony Research \\
  \texttt{bojhuang@gmail.com} \\
  \And
  Jinzhuo Wang \\
  Peking University \\
  \texttt{wangjinzhuo@pku.edu.cn} \\
  \And
  Qiang Fu \\
  Microsoft \\
  \texttt{qifu@microsoft.com} \\
  \And
  Shi Han \\
  Microsoft \\
  \texttt{shihan@microsoft.com} \\
  \And
  Dongmei Zhang \\
  Microsoft \\
  \texttt{dongmeiz@microsoft.com} \\
}

\begin{document}

\maketitle

\begin{abstract}
Supervised Causal Learning (SCL) has shown promise in causal discovery by framing it as a supervised learning problem. However, it suffers from significant out-of-distribution generalization challenges. We reveal three limitations of previous SCL practices: a significant performance gap between synthetic benchmarks and real-world data, fragility to distribution shifts, and failure in compositional generalization, collectively questioning its real-world applicability. To address this, we propose Test-Time Training for Supervised Causal Learning (TTT-SCL), a novel framework that dynamically generates training sets explicitly aligned with any specific test instance. We demonstrate the correlation between TTT-SCL and score-based methods, and design an efficient module for generating training sets based on the classic scoring function. Experiments on synthetic benchmarks, pseudo-real and real-world datasets demonstrate that TTT-SCL significantly outperforms existing SCL and traditional causal discovery methods.
\end{abstract}

\section{Introduction}
\label{sec:intro}
\begin{figure}[t]
\centering
\includegraphics[width=\linewidth]{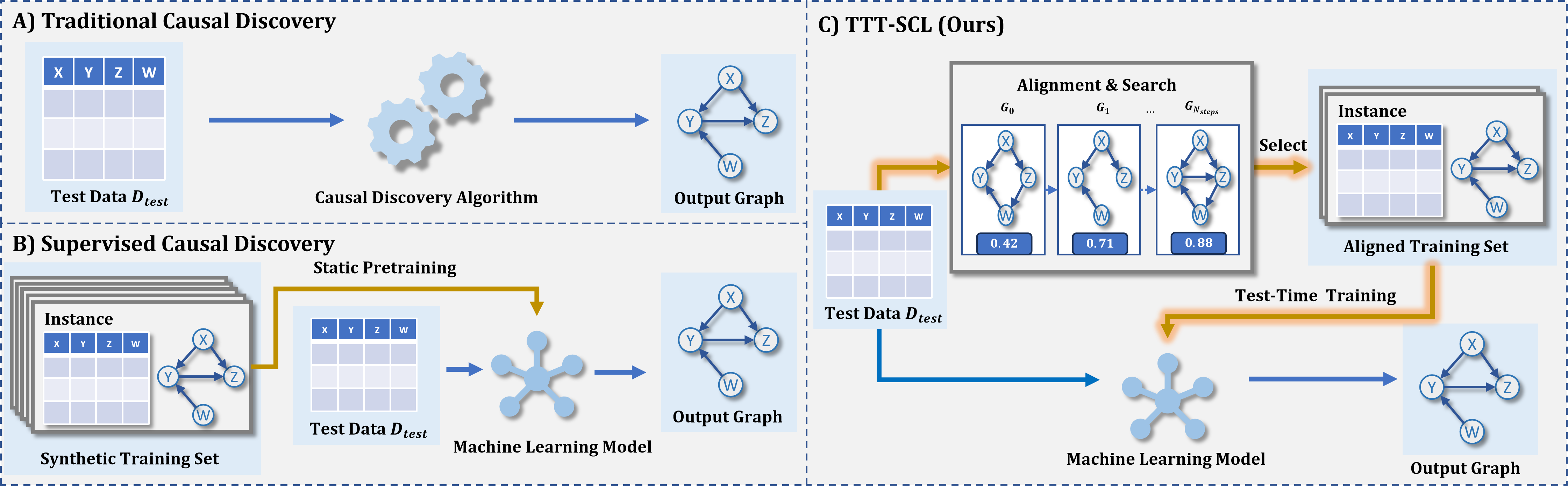}
    \caption{{Test time training supervised causal learning compare with causal discovery (unsupervised causal learning) and supervised causal learning.}}
    \label{framework}
\end{figure}

Causal discovery aims to infer causal relationships from observational data  ~\citep{pearl2009causality, 031spirtes2000causation}. While traditionally approached as an unsupervised problem (Fig. \ref{framework} (A)), Supervised Causal Learning (SCL) has recently emerged as a promising alternative ~\citep{188dai2023ml4c, 189lorch2022amortized, 191ke2022learning, zhang2025learning}. SCL treats causal discovery as a supervised learning task: a model is trained on a set of synthetic causal instances, each comprising a causal graph and its corresponding sampled dataset, and learns to map datasets to their underlying causal graphs (Fig. \ref{framework} (B)).

A pivotal factor for SCL's success is the design of its training set. What properties should these synthetic training sets possess to ensure the model performs well on a real-world, unseen test instance? Two complementary principles guide this design: \textbf{diversity} and \textbf{concentration}. Diversity seeks broad coverage of possible causal models by varying key components such as graph structures, mechanisms, and noise characteristics, thereby encouraging generalization. Concentration, in contrast, aims to align the training set closely with the specific characteristics of the test domain of interest.

Current SCL methods heavily prioritize diversity, pre-training on large, static training sets generated from wide-ranging synthetic distributions ~\citep{189lorch2022amortized, 191ke2022learning}. However, we demonstrate that this paradigm suffers from critical limitations. Through experiments, we observe a notable generalization gap where strong performance on synthetic benchmarks fails to translate to real‑world data, thereby questioning the practical utility of existing SCL approaches. Within the synthetic domain, we further find that these models are fragile under distribution shifts, showing significant performance degradation when the test instance differs from the training set in graph structure, causal mechanism, or noise distribution. More fundamentally, they fail to generalize compositionally; even when trained on all individual components, they cannot handle novel combinations of these components.

These findings motivate a paradigm shift from diversity to concentration. We argue that robust generalization requires moving beyond a single, fixed training set and instead dynamically adapting to each test instance. To this end, we introduce a Test-Time Training for Supervised Causal Learning (TTT-SCL) framework. The core idea is that for a given test dataset, we dynamically generate a new, customized training set that is explicitly aligned with its distribution, train a specialized SCL model, and utilize this model to infer the causal graph (Fig. \ref{framework} (C)). 

Meanwhile, we established a connection between TTT-SCL and score-based methods, under which score-based methods can be considered a special case of TTT-SCL. Based on this, we established a theoretical basis for TTT-SCL and designed a module for efficiently generating training sets. Experiments on synthetic, pseudo-real, and real-world data show that TTT-SCL significantly outperforms existing SCL and traditional causal discovery methods.

Our main contributions are:
\begin{itemize}[left=0pt]
    \item We reveal three limitations of static SCL pre‑training: a significant performance gap between synthetic benchmarks and real‑world data, fragility to distribution shifts, and failure in compositional generalization, collectively questioning its real‑world applicability.
    \item We introduce the TTT-SCL framework, enabling dynamic generation of aligned training sets at test time. We demonstrate that the score-based method can be regarded as a special case of TTT-SCL, thereby establishing a theoretical basis for TTT-SCL and designing a module for efficiently generating training data.
    \item Our experiments confirme that TTT-SCL achieves good performance across both synthetic, pseudo-real and real-world datasets.
\end{itemize}

\section{Background}
\label{sec:framework}

We begin by formalizing the core components of causal learning. A Structural Causal Model (SCM) consists of three key elements: causal graph, causal mechanisms, and noise distributions \citep{pearl2009causality, peters2017elements}. Specifically:

\begin{itemize}[left=0pt]
    \item Causal Graph: Let $G = (V, E)$ be a Directed Acyclic Graph (DAG) with vertex set $V = \{X_1, \dots, X_d\}$ and edge set $E \subseteq V \times V$, where $d$ is the number of variables. The adjacency matrix $A \in \{0,1\}^{d \times d}$ encodes edge relationships where $A_{ij} = 1$ iff $X_i \to X_j \in E$.

    \item Causal mechanisms and noise: Each variable $X_i$ is generated by a causal mechanism and exogenous noise, following the Structural Causal Model (SCM) framework \citep{pearl2009causality}. The data-generating process is characterized by the structural equations:
    \begin{equation}
        X_i := f_i(\mathbf{Pa}_G(X_i), \varepsilon_i),
    \end{equation}
    where $\mathbf{Pa}_G(X_i)$ denotes parents of $X_i$ in $G$, $f_i: \mathbb{R}^{|\mathbf{Pa}_G(X_i)|} \times \mathcal{E}_i \to \mathbb{R}$ is the causal mechanism, and $\varepsilon_i \sim P_{\varepsilon_i}$ is exogenous noise from distribution $P_{\varepsilon_i}$. The full SCM is thus characterized by the tuple $(G, \{f_i\}_{i=1}^d, \{\varepsilon_i\}_{i=1}^d)$, which comprehensively captures the causal structure, functional relationships, and exogenous noise.
\end{itemize}

In supervised causal learning, we work with \textbf{causal instances}. A causal instance is defined by a graph $G$ and a dataset $D$ containing $n$ observations $\{\mathbf{x}^{(1)}, \dots, \mathbf{x}^{(n)}\} \in \mathbb{R}^{n \times d}$, generated from the SCM $(G, \{f_i\}_{i=1}^d, \{\varepsilon_i\}_{i=1}^d)$. 
The \textbf{training set} comprises $K$ such instances, denoted as $\{(D^k_{train}, G^k_{train})\}_{k=1}^K$, where each $D^k_{train}$ is generated from its corresponding $G^k_{train}$. Similarly, at test time, we are given a single \textbf{test instance} $(D_{test}, G_{test})$, where $D_{test}$ is observed but $G_{test}$ is unknown. To avoid notation clutter, we adopt the following conventions: indices $i, j$ refer to variable/node indices within a graph, and subscripts ``train" and ``test" distinguish between training and test entities.

Causal discovery aims to estimate the causal graph $G_{test}$ from $D_{test}$ using a model or algorithm $M$. Supervised causal learning (SCL) frames this as a supervised learning problem, where a model (typically a neural network) is trained on synthetic causal instances to learn a mapping from observational data to graph structures. Formally, the SCL objective is to learn:
\begin{equation}
\mathcal{M}: \mathbb{R}^{n \times d} \to \{0,1\}^{d \times d},
\end{equation}
which maps an input data matrix (e.g., $D_{test}$) to an output adjacency matrix (representing $G_{test}$). The model is trained on synthetic pairs $\{(D^k_{train}, G^k_{train})\}_{k=1}^K$.

Previous SCL methods rely on training with synthetic data, where the generative distribution is explicitly controlled along three dimensions consistent with the SCM framework: graph structure, causal mechanisms, and noise distributions \citep{189lorch2022amortized, 191ke2022learning, froehlich2024graph}.

\section{Out-of-distribution Challenges for SCL}
\label{sec:ood}

Supervised causal learning (SCL) relies almost entirely on synthetic data for training because real-world causal graphs are rarely available, which naturally raises the concern that performance measured on synthetic benchmarks may not faithfully reflect performance on real-world data. Our results on the Sachs dataset confirm this concern, as competitive performance on synthetic benchmarks does not translate into comparable accuracy on real-world data (\textbf{Issue 1}). A closer look within the synthetic domain further reveals why static pre-training breaks down. The training distribution is jointly characterized by three dimensions, namely graph structure, causal mechanism, and noise distribution, and a mismatch in any single dimension between training and test already causes substantial performance degradation, exposing the fragility of SCL under distribution shifts (\textbf{Issue 2}). More fundamentally, even when the model has encountered all individual settings across these dimensions during training, it still fails on unseen combinations of them, demonstrating a clear failure of compositional generalization (\textbf{Issue 3}). The experiments below illustrate these three limitations.
    
\subsection{Experiment Setup}

\textbf{Datasets.} To comprehensively evaluate generalization, we use both synthetic benchmarks, pseudo-real and real-world datasets.
\begin{itemize}[left=0pt]
    \item \textbf{Synthetic data:} We generate test instances from a factorial combination of mechanism, graph and noise distribution. We use three mechanism classes: Linear, Random Fourier Features (RFF) \citep{rahimi2007random}, and Chebyshev polynomials \citep{froehlich2024graph}. We use two random graph models: Erdos-Renyi (ER) and Scale-Free (SF) ~\citep{gilbert1959random, barabasi2009scale}. Gaussian noise is used for RFF and Chebyshev mechanisms, while Uniform noise is used for Linear mechanisms to ensure identifiability. This yields six test settings: RFF\_G\_ER, RFF\_G\_SF, Linear\_U\_ER, Linear\_U\_SF, Chebyshev\_G\_ER, and Chebyshev\_G\_SF.

    \item \textbf{Pseudo-real data}: We also incorporate pseudo-real datasets generated by the SynTReN generator \citep{van2006syntren}. This generator is specifically designed to simulate synthetic transcriptional regulatory networks with biologically plausible structures and parameters, producing gene expression data that closely resembles experimental microarray data.
    
    \item \textbf{Real-world data:} We use the Sachs dataset ~\citep{sachs2005}, a well-established benchmark in causal discovery. It contains 853 measurements of 11 proteins and a consensus causal graph derived from biological knowledge.

\end{itemize}

\textbf{Model \& Training setting.}: We mainly use AVICI as the model backbone \citep{189lorch2022amortized}, a DNN-based architecture which is currently widely followed by the community and open-sourced. To assess different generalization aspects, we compare several training settings, more detailed configurations can be found in Appendix \ref{config detail}:

\begin{itemize}[left=0pt]
    \item \textbf{i.i.d}: The training data and test data are exactly the same distribution.

    \item \textbf{Graph/Noise/Mechanism shift}: The mechanism/graph/noise of the training data is different from that of the test data, but the other two distributions are the same.

    \item \textbf{Component-mixed}: This training setup contains all individual components (mechanisms, graph types, noise distributions) seen in isolation during training, but crucially excludes the specific combinations present in the test instances. This tests whether the model can perform \textit{compositional generalization} by recombining learned components, rather than merely memorizing training configurations.
    
    \item \textbf{AVICI (scm-v0)}: This model was trained on SCM data simulated from a large variety of graph models with up to 100 nodes, both linear and nonlinear causal mechanisms, and homogeneous and heterogeneous additive noise from Gaussian, Laplace, and Cauchy distributions. It can be considered one of the strongest model of open source under the SCL paradigm \footnote{https://github.com/larslorch/avici}.

\end{itemize} 

\begin{table}[htbp]
    \centering
    \caption{Divergent generalization patterns. Strong synthetic performance does not guarantee effectiveness on real-world data. Results are presented as AUROC \color{gray}{\scalebox{0.85}{(standard deviation)}}.} 
    \label{tab:divergent_patterns}  
    \begin{tabular}{l l l l l l}
        \toprule
        & RFF\_G & Linear\_U & Chebyshev\_G & Sachs & Syntren\\
        \midrule
        PC & 61.1 \color{gray}{\scalebox{0.85}{(4.9)}} & 60.9 \color{gray}{\scalebox{0.85}{(4.7)}} & 59.8 \color{gray}{\scalebox{0.85}{(6.6)}} & 67.1 & 58.1 \\
        AVICI (scm-v0) & 97.8 \color{gray}{\scalebox{0.85}{(1.3)}} & 75.6 \color{gray}{\scalebox{0.85}{(13.8)}} & 81.7 \color{gray}{\scalebox{0.85}{(10.5)}} & 62.3 & 65.4\\
        \bottomrule
    \end{tabular}
\end{table}

\subsection{Limitations of Current SCL Paradigms}

Our experimental results validate the three issues outlined above, collectively exposing the limitations of static pre-training in SCL.

\textbf{Issue 1}. The results in Table \ref{tab:divergent_patterns} question the value of synthetic benchmarks by demonstrating that strong synthetic performance fails to guarantee effectiveness on real-world data. Here, we merge the dimensions of the graph and analyze more from the perspective of the mechanism. While AVICI (scm-v0) excels on synthetic data similar to its training distribution (e.g., RFF\_G, 97.8), its performance collapses on the real-world Sachs dataset (62.3). In contrast, traditional methods like PC maintain consistent, albeit lower, performance across domains. This divergence reveals that SCL models overfit to the artifacts of their synthetic training set, lacking the cross-domain consistency required for real-world applicability.

\begin{figure}
 \centering
 \includegraphics[width=0.7\linewidth]{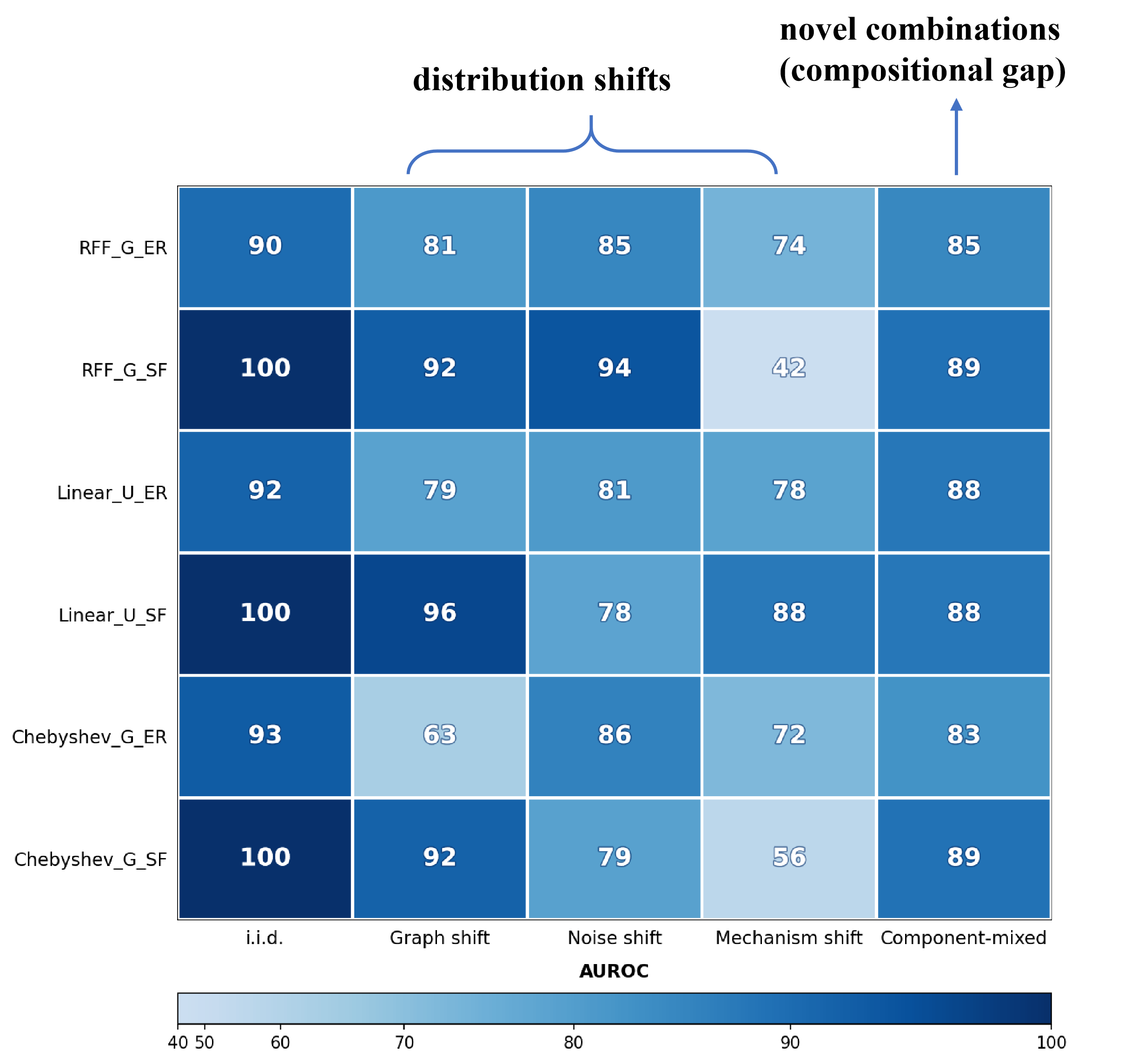}
 \caption{Two limitations of static SCL: fragility to distribution shifts and failure in compositional generalization. Each column represents a test setting, and each row shows the specific distribution of test data. The specific training distribution for the 5 test settings can be found in Appendix \ref{config detail}.}
  \label{fig:ood}
\end{figure}

\textbf{Issue 2}. The results in Fig.~\ref{fig:ood} demonstrate that distribution shifts across all three dimensions (graph structure, causal mechanism, and noise distribution) significantly degrade SCL performance. Models struggle when the test-time graph structure (``Graph shift" compared to ``i.i.d"), causal mechanism (``Mechanism shift" compared to ``i.i.d"), or noise distribution (``Noise shift" compared to ``i.i.d") differs categorically from those seen during training. While performance drops are observed in all cases, ``Mechanism shifts" emerge as particularly damaging, underscoring the profound impact of the underlying mechanism functional form on model generalization.

\textbf{Issue 3}. Even when trained on data containing all individual components, the model still exhibits performance drop on unseen combinations of these components, as seen when comparing ``Component-mixed'' to ``i.i.d'' in Fig.~\ref{fig:ood}. This compositional failure indicates that SCL models memorize specific $(G, f, \varepsilon)$ configurations rather than learning a modular understanding of causal factors.


In summary, the dual failure of fragility under distribution shifts and inconsistency across domains fundamentally undermines the static pre-training paradigm. These limitations are not artifacts of a specific architecture, as validated by consistent failure patterns using the SiCL backbone (Appendix~\ref{other backbone}). The results compellingly argue that robust causal discovery requires a shift from static, diversity-seeking pre-training to dynamic, test-time adaptation.

\section{Test-Time Training for Supervised Causal Learning}
\label{sec:tttscl}

Given an observed test dataset $D_{{test}}$, TTT-SCL infers its underlying causal graph by training a supervised model at test time on data that are causally aligned with $D_{{test}}$, as illustrated in Fig.~\ref{framework}(C). The workflow proceeds as follows: a training set tailored to $D_{{test}}$ is first generated under the guidance of a score function, and a supervised causal learning model is then trained on this set and applied to $D_{{test}}$ for inference. In our work, the design and acquisition of the test-time training set constitutes the core technical contribution, while the supervised causal learning module can leverage any existing architecture. Therefore, we focus on the generation of the training set in the following, and use the score as a bridge to analyze the relationship between TTT-SCL and classical score-based methods.

\subsection{Causally Aligned Training Set Generation}
\label{sec:training set generation}

To generate a training set that is causally aligned with $D_{{test}}$, we need a principled measure of alignment quality, which motivates our adoption of a score function that evaluates how well a candidate graph $G$ fits the observed data $D_{test}$. Formally, we adopt a score
\begin{equation}
    \mathrm{score}(G, D_{test}) = \mathrm{AD}(G, D_{test}) - \|\mathbf{A}_G\|_0,
    \label{eq:score}
\end{equation}
where the alignment-of-distribution (AD) term is defined as the average log-likelihood of the variables under the mechanisms induced by $G$:

\begin{equation}
    \mathrm{AD}(G, D_{{test}}) \coloneqq \frac{1}{d}\sum_{i=1}^{d} \frac{1}{n}\sum_{t=1}^{n} \log p\bigl(x_i^{(t)} \mid f_i^{G}(\mathbf{x}_{\mathrm{Pa}_G(i)}^{(t)})\bigr).
    \label{eq:ad}
\end{equation}

Here $f_i^{G}$ denotes the causal mechanism of variable $X_i$ given its parents in $G$, fitted on $D_{{test}}$ by regressing $X_i$ on $\mathbf{Pa}_G(X_i)$ via Structure-Induced Mechanism (SIM). For a candidate graph, SIM estimates each variable's functional relationship with its parents and the residual noise from the test data, enabling the likelihood evaluation above as well as the generation of causally aligned synthetic datasets described later (see Appendix~\ref{app:sim} for details). The term $\| \mathbf{A}_G\|_0$ counts the number of edges and penalizes graph complexity, in the spirit of classical information criteria. Guided by this score, we can obtain a collection of high-scoring graphs. 

However, finding these graphs is difficult as exhaustively searching the entire DAG space is intractable. To solve this problem, we propose an informed construction approach which combines good initializations with guided refinement. Concretely, it proceeds in three stages:

\begin{enumerate}[left=0pt]
    \item \textbf{Seed Initialization.} We start from an initial graph $G^{{0}}$ obtained by applying a traditional causal discovery method (e.g., PC, NOTEARS) on $D_{{test}}$. Note that the initial graph can also be a random DAG, while using a better initial graph provides a useful prior and accelerates the whole process.

    \item \textbf{Iterative Stochastic Graph Refinement.} From the current graph $G_{curr}$, we iteratively propose local modifications to the graph and obtain $G_{cand}$. The possible modifications include single-edge additions, deletions, and reversals do not break the acyclicity. After that, the candidate $G_{cand}$ is evaluated using the joint score function in Eq. (\ref{eq:score}) and is accepted with probability proportional $\alpha = \min\left[1, \frac{score\left(G_{cand}, D_{test}\right)}{score\left(G_{curr}, D_{test}\right)}\right]$. This stochastic refinement process is repeated iteratively and yields multiple graphs during the last multiple iterations.

    \item \textbf{Training Data Generation.} For each graph $G^k_{train}$, we regress its mechanism with $D_{test}$ via SIM, and then forward-sample a synthetic dataset $D^k_{train}$ based on the regressed mechanism. All causal instances ($D^k_{train}$, $G^k_{train}$) then assemble the final training sets $\{(D^k_{train},G^k_{train})\}_{k=1}^K$.
\end{enumerate}

The complete TTT-SCL workflow is formalized in Algorithm ~\ref{alg:ttt-scl}. By combining score and stochastic refinement, this approach realizes an efficient and directed method for searching the graph space at test time. All hyperparameter details are described in the Appendix \ref{config detail}. A diagram illustrating the workflow of training sets generation can be found in Appendix ~\ref{Workflow of training sets generation}.

\begin{algorithm}[t]
\caption{Test-Time Training for Supervised Causal Learning (TTT-SCL)}
\label{alg:ttt-scl}
\begin{algorithmic}[1]
\Require 
    \Statex Test dataset $D_{\text{test}} \in \mathbb{R}^{n \times d}$, Number of refinement steps $N_{\text{steps}}$, number of training graphs $K$.
\Ensure Predicted causal graph ${G}_{\text{pred}} \in \{0,1\}^{d \times d}$.
\State $G^0 \gets \text{InitialGraph}(D_{\text{test}})$ \Comment{e.g., NOTEARS, or a random DAG}
\State Initialize graph list $\mathcal{G} \gets []$
\State $G_{\text{curr}} \gets G^0$
\For{$t = 1$ \textbf{to} $N_{\text{steps}}$}
    \State $G_{\text{cand}} \gets \text{Local}(G_{\text{curr}})$ 
        \Comment{Add, delete, or reverse an edge while preserving acyclicity}
    \State $s_{\text{curr}} \gets \text{Score}(G_{\text{curr}}, D_{\text{test}})$
    \State $s_{\text{cand}} \gets \text{Score}(G_{\text{cand}}, D_{\text{test}})$
    \State $\alpha \gets \min\!\big(1,\, s_{\text{cand}} / s_{\text{curr}}\big)$
    \If{$\text{random}(0,1) < \alpha$}
        \State $G_{\text{curr}} \gets G_{\text{cand}}$
    \EndIf
    \If{$t > N_{\text{steps}} - K$}
        \State $\mathcal{G}.\text{append}(G_{\text{curr}})$ \Comment{Collect the last $K$ graphs}
    \EndIf
\EndFor
\State Initialize training set $\mathcal{T} \gets \emptyset$
\For{each $G^k_{train} \in \mathcal{G}$}
    \State $\{f_i, \hat{P}_{\epsilon_i}\}_{i=1}^d \gets \text{SIM}(G^k_{train}, D_{\text{test}})$
        \Comment{Regress mechanisms using $D_{\text{test}}$ via SIM}
    \State $D^k_{train} \gets \text{Sample}(G^k_{train}, \{f_i, \hat{P}_{\epsilon_i}\}_{i=1}^d, n)$
    \State $\mathcal{T} \gets \mathcal{T} \cup \{(D^k_{train}, G^k_{train})\}$
\EndFor
\State Train a supervised causal learning model $\mathcal{M}$ on $\mathcal{T}$ \Comment{e.g., AVICI}
\State ${G}_{\text{pred}} \gets \mathcal{M}(D_{\text{test}})$
\State \Return ${G}_{\text{pred}}$
\end{algorithmic}
\end{algorithm}

\subsection{Relation to Classical Score-Based Causal Discovery}

Classical score-based causal discovery methods \citep{054chickering2002optimal,060zheng2018dags,Lachapelle2020Gradient-Based} find an optimal graph based on the score, while TTT-SCL gives the score a richer meaning and more flexible usage. \textbf{First}, we move from selecting a single optimal graph to building a set of candidate graphs, thus replacing a point estimate with a collection that is more informative for learning. Under finite samples, multiple graphs often more plausible due to statistical uncertainty and score approximation error. \textbf{Second}, we do not simply select the top $K$ graphs by score value. Instead, the set of candidate graphs preserves sturtural diversity, effectively covering a neighborhood of plausible structures around the test instance's underlying causal graph, as described in Sec ~\ref{sec:training set generation}. This neighborhood captures the intrinsic uncertainty that arises from finite samples and score imprecision. \textbf{Third}, we do not use the score as a final decision rule; rather, we train a supervised model on these instance‑specific training sets to further optimize predictive accuracy. This learned model can combine evidence across the entire candidate set and exploit patterns that are not fully captured by the scalar score alone. In Sec ~\ref{sec:relationship}, our experimental results confirm the performance improvement brought about by supervised learning.

\paragraph{Further, score‑based methods can be regarded as a special case of TTT‑SCL.}Given a test dataset $D_{{test}}$ and a set of training pairs $\{(D_{{train}}^k, G_{{train}}^k)\}_{k=1}^{K}$, let $s_k \coloneqq \mathrm{score}(G_{{train}}^k; D_{{test}})$ and $k^\star = \arg\max_k s_k$.
A score‑based method returns $G_{k^\star}$.
If we instantiate TTT‑SCL with the same training set and choose the supervised predictor $f$ as the $1$-nearest-neighbor classifier based on the score values, then the TTT‑SCL output on $D_{{test}}$ coincides exactly with the score‑based output: $f(D_{{test}}) = G_{k^\star}$. Thus, classical score‑based selection corresponds to a particular, restricted configuration of TTT‑SCL.

Because TTT‑SCL generalizes the score‑based principle, it inherits the theoretical guarantees of the underlying score. Under standard identifiability assumptions on the data‑generating process, such as linear non‑Gaussian models, additive noise models, and other settings where the causal graph is uniquely determined by the observational distribution \citep{055shimizu2006linear, 056hoyer2008nonlinear, zhang2009pnl}, maximizing a properly defined score recovers the true causal graph \citep{Lachapelle2020Gradient-Based}. Since our score instantiates the same likelihood-plus-sparsity structure, high‑scoring graphs concentrate around the true causal structure, and the learner trained on data generated by these graphs is at least as reliable as the single highest‑scoring graph. Moreover, the supervised model can outperform score‑based selection in finite samples by exploiting the richer information contained in the collection of candidates. This flexibility also allows TTT‑SCL to accommodate different score designs tailored to specific model classes without changing the overall framework.

\section{Experiments}
\label{sec:experiment}

In this section, we compare the performance of TTT-SCL with multiple baseline methods on various synthetic data, pseudo-real data and real data. These datasets are consistent with the content of Section 3.1.

\textbf{Baselines:} We compare against traditional causal discovery methods PC ~\citep{031spirtes2000causation}, GES ~\citep{054chickering2002optimal}, NOTEARS ~\citep{060zheng2018dags}, {RESIT ~\citep{peters2014causal}, SCORE ~\citep{rolland2022score}, NoGAM ~\citep{montagna2023causal} and AVICI ~\citep{189lorch2022amortized}, a DNN-based SCL method which is currently widely followed by the community and open source. We use the open-source pre-trained AVICI (scm-v0) model, which is trained on a vast mixture of synthetic data and represents the strongest publicly available SCL baseline.

\textbf{Our Method (TTT-SCL):} For our TTT-SCL approach, we set the number of dynamically generated training graphs to $K=200$. The number of variables $d$ is 10 for synthetic data, 11 for Sachs and 20 for Syntren. The observation $n$ for each generated dataset matches that of the test data. We evaluate two variants of our method: TTT-SCL (random) which initializes the seed graph with a random DAG, and TTT-SCL (Notears) which uses a graph estimated from $D_{test}$ by the NOTEARS algorithm as a smarter starting point. 

\textbf{Evaluation metrics:} We use multiple metrics to evaluate the predicted graphs, including Area Under the Receiver Operating Curve (AUROC), Area Under the Precision-Recall Curve (AUPRC), F1 score and Accuracy (ACC). In the main text, we primarily report \textbf{AUROC} for edge prediction to succinctly explore the impact of training data quality on model performance.

\subsection{The Performance of TTT-SCL}

\begin{table}[htbp]
    \centering
    \caption{TTT-SCL performance on synthetic, real and pseudo-real datasets. Results are presented as AUROC \color{gray}{\scalebox{0.85}{(standard deviation)}}.} 
    \label{tab:performance_TTT-SCL}
    \begin{tabular}{l l l l l l}
        \toprule
         & RFF\_G & Linear\_U & Chebyshev\_G & Sachs & Syntren \\
        \midrule
        PC & 61.1 \color{gray}{\scalebox{0.85}{(4.9)}} & 60.9 \color{gray}{\scalebox{0.85}{(4.7)}} & 59.8 \color{gray}{\scalebox{0.85}{(6.6)}} & 67.1 & 58.1 \\
        GES & 66.0 \color{gray}{\scalebox{0.85}{(10.6)}} & 69.0 \color{gray}{\scalebox{0.85}{(10.8)}} & 59.6 \color{gray}{\scalebox{0.85}{(5.9)}} & 61.8 & 36.8 \\
        Notears & 80.5 \color{gray}{\scalebox{0.85}{(4.0)}} & 82.0 \color{gray}{\scalebox{0.85}{(4.6)}} & 52.2 \color{gray}{\scalebox{0.85}{(3.5)}} & 61.8 & 49.8 \\
        {RESIT} &  {54.3} \color{gray}{\scalebox{0.85}{(5.4)}} & {54.1} \color{gray}{\scalebox{0.85}{(5.2)}} & {49.8} \color{gray}{\scalebox{0.85}{(4.7)}} & {62.3} & {64.6} \\
        {SCORE} & {86.9} \color{gray}{\scalebox{0.85}{(3.2)}} & {82.2} \color{gray}{\scalebox{0.85}{(18.7)}} & {69.2} \color{gray}{\scalebox{0.85}{(7.6)}} & {64.9} & {41.0} \\
        {NoGAM} & {87.6} \color{gray}{\scalebox{0.85}{(2.9)}} & {79.2} \color{gray}{\scalebox{0.85}{(18.6)}} & {72.3} \color{gray}{\scalebox{0.85}{(6.4)}} & {64.9} & {41.0} \\
        AVICI (scm-v0) & \textbf{97.8} \color{gray}{\scalebox{0.85}{(1.3)}} & 75.6 \color{gray}{\scalebox{0.85}{(13.8)}} & 81.7 \color{gray}{\scalebox{0.85}{(10.5)}} & 62.3 & 65.4 \\
        TTT-SCL (random) & 88.4 \color{gray}{\scalebox{0.85}{(7.0)}} & 82.3 \color{gray}{\scalebox{0.85}{(7.0)}} & 79.6 \color{gray}{\scalebox{0.85}{(6.7)}} & 58.6 & 72.0 \\
        TTT-SCL (Notears) & 91.8 \color{gray}{\scalebox{0.85}{(3.1)}} & \textbf{86.3} \color{gray}{\scalebox{0.85}{(4.4)}} & \textbf{83.0} \color{gray}{\scalebox{0.85}{(8.7)}} & \textbf{78.9} & \textbf{80.1} \\
        \bottomrule
    \end{tabular}
    
\end{table}

The results are summarized in Table~\ref{tab:performance_TTT-SCL}. Overall, TTT-SCL demonstrates robust and highly competitive performance. The pre-trained AVICI (scm-v0) model achieves optimal performance on the RFF\_G datasets, as it was explicitly trained on this distribution. TTT-SCL's performance on RFF\_G is slightly lower but remains strong, indicating its ability to approximate even in-distribution performance without prior exposure. Crucially, TTT-SCL achieves state-of-the-art performance on all other datasets, including Linear\_U, Chebyshev\_G, real-world Sachs, and pseudo-real Syntren dataset. This confirms that TTT-SCL excels in the most challenging and realistic scenarios involving distribution shifts, where static pre-training fails. 

These conclusions hold consistently across multiple evaluation metrics, as demonstrated in Appendix~\ref{other metrics}, where TTT-SCL maintains superior performance in ACC, F1-score, and AUPRC under various distribution shifts. Appendix~\ref{app:bnlearn_experiments} presents results on four additional established benchmark causal graphs from the bnlearn repository (Asia, Cancer, Earthquake, and Survey), demonstrating TTT-SCL's robust performance across diverse real-world causal structures. The complexity analysis and runtime variation with the number of nodes are detailed in Appendix \ref{app:runtime}. 

TTT-SCL (Notears) consistently outperforms TTT-SCL (random), indicating that a reasonable initial graph provides a useful prior and accelerates the refinement process. At the same time, the competitive performance of TTT-SCL (random) confirms the inherent robustness of our stochastic search. We further note that the gap between the two variants can be substantially narrowed by increasing the search budget, suggesting that sensitivity to the seed graph primarily reflects computational efficiency rather than a fundamental limitation. A detailed analysis is given in Appendix~\ref{app:seed}. 

\subsection{Ablation Study}
We design experiments to empirically validate how these two components contribute to the quality of the generated training data. We first ablate the sparsity term in the optimization objective to isolate its effect. We compare the full TTT-SCL (Notears) method against a variant, TTT-SCL (Notears-s), where the sparsity penalty is removed, thus optimizing for AD alone. Results in Table ~\ref{tab:ablation_sparsity} show that removing the sparsity term leads to a consistent and significant performance drop across all test settings. 

\begin{table}[h!]
\centering
\caption{Ablation experiment of sparsity. Results are presented as AUROC.}
\label{tab:ablation_sparsity}
\begin{tabular}{llllll}
\toprule
 & RFF\_G & Linear\_U & Chebyshev\_G & Sachs & Syntren \\ \midrule
TTT-SCL (Notears) & 91.8 \color{gray}{\scalebox{0.85}{(3.1)}} & 86.3 \color{gray}{\scalebox{0.85}{(4.4)}} & 83.0 \color{gray}{\scalebox{0.85}{(8.7)}} & 78.9 & 80.1 \\ 
TTT-SCL (Notears-s) & 86.8 \color{gray}{\scalebox{0.85}{(2.9)}} & 84.3 \color{gray}{\scalebox{0.85}{(7.9)}} & 69.7 \color{gray}{\scalebox{0.85}{(12.4)}} & 63.5 & 76.1 \\ \bottomrule
\end{tabular}
\end{table}

To further demonstrate the effectiveness of AD and the necessity of sparsity, the AD, sparsity, score of the training data obtained by different methods under different test data, as well as the AUROC on the test data were recorded in Appendix \ref{Ablation study}. The results show that both AD and sparsity are indispensable and important elements, and they have certain indicative significance for performance.

\subsection{Relationship with Score-based Methods}
\label{sec:relationship}

To clearly distinguish our approach from classical score-based methods, we provide a detailed stage-wise analysis comparing three key outputs: the seed graph, the highest-scoring graph found during TTT-SCL's search, and the final SCL prediction. The consistent performance improvement across these stages demonstrates the added value of the supervised learning phase. Specifically, we conducted experiments comparing three different outputs across four test domains (RFF, Linear, Chebyshev, and Sachs) for detailed analysis: 

\begin{enumerate}[left=0pt]
    \item \textbf{Seed graph}: Initial graph from proxy methods
    \item \textbf{Highest-score graph}: Highest-score graph found during TTT-SCL's stochastic refinement
    \item \textbf{Final output}: Graph predicted by the SCL model trained on TTT-SCL-generated data
\end{enumerate}

\begin{table}[h!]
    \caption{Performance comparison (AUROC) across different stages of TTT-SCL.}
    \label{tab:TTT-SCL_stages}
    \centering
    \small %
    \begin{tabular}{lcccc}
        \toprule
        & RFF\_G    & Linear\_U & Chebyshev\_G & Sachs  \\
        \midrule
        Seed graph        & 80.5  & 82.0  & 52.2     & 61.8  \\
        Highest-score graph in TTT-SCL search & 88.9  & 80.1  & 75.8     & 66.6  \\
        Final graph from trained SCL model   & \textbf{91.8}  & \textbf{86.3}  & \textbf{83.0}     & \textbf{78.9}  \\
        \bottomrule
    \end{tabular}

\end{table}

The results in Table \ref{tab:TTT-SCL_stages} clearly demonstrate the two-stage improvement of our approach:

\begin{itemize}[left=0pt]
    \item \textbf{1$\rightarrow$2 (Search Improvement)}: The higher AUROC of the highest-score graph compared to the seed graph shows that TTT-SCL's stochastic refinement effectively improves graph quality through distributional alignment.
    
    \item \textbf{2$\rightarrow$3 (Learning Improvement)}: The consistent and substantial performance gain of the final SCL output over the highest-score graph demonstrates the crucial advantage of our approach. While score-based methods would stop at the highest-scoring graph, TTT-SCL uses this graph to generate training data that enables an SCL model to learn more accurate causal relationships.
\end{itemize}

This two-stage process, where we optimize for training data quality rather than directly for the final graph, constitutes the fundamental distinction between TTT-SCL and classical score-based causal discovery.

\section{Related Work}
\label{sec:related work}

Causal discovery has traditionally been approached as an unsupervised problem, with constraint-based methods (e.g., PC, FCI \citep{031spirtes2000causation}), function-based methods (e.g., LiNGAM \citep{055shimizu2006linear}, ANM \citep{056hoyer2008nonlinear}), and score-based methods (e.g., GES \citep{054chickering2002optimal}, NOTEARS \citep{060zheng2018dags}, DAG-GNN \citep{085yu2019dag}, GraN-DAG \citep{Lachapelle2020Gradient-Based}) inferring causal graphs directly from observational data. While principled, these methods often suffer from high sample complexity, sensitivity to faithfulness violations, and limited scalability.

Supervised Causal Learning (SCL) has recently emerged as a promising alternative that frames causal discovery as a supervised learning problem \citep{188dai2023ml4c, 189lorch2022amortized, 191ke2022learning}. Prior SCL methods employ diverse architectures, including cascade classifiers for sequential independence testing \citep{187ma2022ml4s}, attention-based transformers operating on data as 3D tensors \citep{189lorch2022amortized, 191ke2022learning, froehlich2024graph}, and pairwise attention networks \citep{zhang2025learning}. Their target outputs also vary, ranging from undirected skeletons \citep{187ma2022ml4s} and local v-structure classifications \citep{188dai2023ml4c} to full directed adjacency matrices \citep{189lorch2022amortized, 191ke2022learning}, with some methods only guaranteeing recovery up to Markov equivalence \citep{zhang2025learning, froehlich2024graph}.

Regarding training data strategy, \cite{montagna2024demystifying} recently investigated SCL generalization challenges but attributed performance drops primarily to unseen individual components and advocated increased pre-training diversity. In contrast, we identify a more fundamental limitation: \textit{compositional generalization} failure, where SCL models fail on novel combinations of seen components, revealing the intractability of exhaustive static pre-training and motivating our TTT-SCL paradigm. 

Finally, while test-time adaptation has shown promise in general machine learning \citep{liang2025comprehensive, sun2020test, wang2020tent, liu2021ttt++, sinha2023test, yuksekgonul2026learningdiscovertesttime}, its application to causal discovery is gaining traction. Concurrently, TICL \citep{chen2026ticl} addresses interventional settings with discrete data via self-augmentation, whereas our TTT-SCL provides a general framework for observational settings with continuous data, generating causally aligned training data at test time.

\section{Conclusion}
In this work, we identified limitations of static SCL paradigms, demonstrating their fragility under distribution shifts, failure in compositional generalization, and poor transfer from synthetic benchmarks to real-world data. To address these out-of-distribution generalization challenges, we introduced TTT-SCL, a paradigm-shifting framework through test-time training of causally-aligned data. TTT-SCL dynamically generates high-quality training data tailored to each test instance, achieving good performance on both synthetic, pseudo-real and real-world datasets. This work not only advances the field of supervised causal learning but also opens new avenues for robust and adaptive causal discovery in real-world settings.

\bibliographystyle{plain}
\bibliography{reference}

@book{031spirtes2000causation,
  title={Causation, prediction, and search},
  author={Spirtes, Peter and Glymour, Clark N and Scheines, Richard},
  year={2000},
  publisher={MIT press}
}

@article{054chickering2002optimal,
  title={Optimal structure identification with greedy search},
  author={Chickering, David Maxwell},
  journal={Journal of machine learning research},
  volume={3},
  number={Nov},
  pages={507--554},
  year={2002}
}

@article{055shimizu2006linear,
  title={A linear non-Gaussian acyclic model for causal discovery.},
  author={Shimizu, Shohei and Hoyer, Patrik O and Hyv{\"a}rinen, Aapo and Kerminen, Antti and Jordan, Michael},
  journal={Journal of Machine Learning Research},
  volume={7},
  number={10},
  year={2006}
}

@inproceedings{056hoyer2008nonlinear,
author = {Hoyer, Patrik O. and Janzing, Dominik and Mooij, Joris and Peters, Jonas and Sch\"{o}lkopf, Bernhard},
title = {Nonlinear causal discovery with additive noise models},
year = {2008},
booktitle = {Proceedings of the 22nd International Conference on Neural Information Processing Systems},
pages = {689–696},
numpages = {8},
location = {Vancouver, British Columbia, Canada},
}

@inproceedings{060zheng2018dags,
author = {Zheng, Xun and Aragam, Bryon and Ravikumar, Pradeep and Xing, Eric P.},
title = {DAGs with NO TEARS: continuous optimization for structure learning},
year = {2018},
booktitle = {Proceedings of the 32nd International Conference on Neural Information Processing Systems},
pages = {9492–9503},
numpages = {12},
location = {Montr\'{e}al, Canada},
}

@inproceedings{187ma2022ml4s,
  title={Ml4s: Learning causal skeleton from vicinal graphs},
  author={Ma, Pingchuan and Ding, Rui and Dai, Haoyue and Jiang, Yuanyuan and Wang, Shuai and Han, Shi and Zhang, Dongmei},
  booktitle={Proceedings of the 28th ACM SIGKDD Conference on Knowledge Discovery and Data Mining},
  pages={1213--1223},
  year={2022}
}

@inproceedings{188dai2023ml4c,
  title={Ml4c: Seeing causality through latent vicinity},
  author={Dai, Haoyue and Ding, Rui and Jiang, Yuanyuan and Han, Shi and Zhang, Dongmei},
  booktitle={Proceedings of the 2023 SIAM International Conference on Data Mining (SDM)},
  pages={226--234},
  year={2023},
  organization={SIAM}
}

@inproceedings{189lorch2022amortized,
author = {Lorch, Lars and Sussex, Scott and Rothfuss, Jonas and Krause, Andreas and Sch\"{o}lkopf, Bernhard},
title = {Amortized inference for causal structure learning},
year = {2022},
booktitle = {Proceedings of the 36th International Conference on Neural Information Processing Systems},
articleno = {952},
pages = {13104 - 13118},
numpages = {15},
location = {New Orleans, LA, USA},
}

@article{191ke2022learning,
  title={Learning to induce causal structure},
  author={Ke, Nan Rosemary and Chiappa, Silvia and Wang, Jane and Goyal, Anirudh and Bornschein, Jorg and Rey, Melanie and Weber, Theophane and Botvinic, Matthew and Mozer, Michael and Rezende, Danilo Jimenez},
  journal={arXiv preprint arXiv:2204.04875},
  year={2022}
}

@inproceedings{froehlich2024graph,
author = {Froehlich, Philipp and Koeppl, Heinz},
title = {Graph structure inference with BAM: neural dependency processing via bilinear attention},
year = {2024},
booktitle = {Proceedings of the 38th International Conference on Neural Information Processing Systems},
articleno = {4093},
pages = {128847-128885},
numpages = {39},
location = {Vancouver, BC, Canada},
}

@inproceedings{
zhang2025learning,
title={Learning Identifiable Structures Avoids Bias in {DNN}-based Supervised Causal Learning},
author={Jiaru Zhang and Rui Ding and Qiang Fu and Huang Bojun and zizhen Deng and Yang Hua and Haibing Guan and Shi Han and Dongmei Zhang},
booktitle={The 28th International Conference on Artificial Intelligence and Statistics},
year={2025},
url={https://openreview.net/forum?id=8JwuE2HhY8}
}

@inproceedings{sun2020test,
  title={Test-time training with self-supervision for generalization under distribution shifts},
  author={Sun, Yu and Wang, Xiaolong and Liu, Zhuang and Miller, John and Efros, Alexei and Hardt, Moritz},
  booktitle={International conference on machine learning},
  pages={9229--9248},
  year={2020},
  organization={PMLR}
}

@article{wang2020tent,
  title={Tent: Fully test-time adaptation by entropy minimization},
  author={Wang, Dequan and Shelhamer, Evan and Liu, Shaoteng and Olshausen, Bruno and Darrell, Trevor},
  journal={arXiv preprint arXiv:2006.10726},
  year={2020}
}

@inproceedings{liu2021ttt++,
author = {Liu, Yuejiang and Kothari, Parth and van Delft, Bastien and Bellot-Gurlet, Baptiste and Mordan, Taylor and Alahi, Alexandre},
title = {TTT++: when does self-supervised test-time training fail or thrive?},
year = {2021},
booktitle = {Proceedings of the 35th International Conference on Neural Information Processing Systems},
articleno = {1669},
pages = {21808-21820},
numpages = {13},
}

@inproceedings{sinha2023test,
  title={Test: Test-time self-training under distribution shift},
  author={Sinha, Samarth and Gehler, Peter and Locatello, Francesco and Schiele, Bernt},
  booktitle={Proceedings of the IEEE/CVF Winter Conference on Applications of Computer Vision},
  pages={2759--2769},
  year={2023}
}

@article{liang2025comprehensive,
  title={A comprehensive survey on test-time adaptation under distribution shifts},
  author={Liang, Jian and He, Ran and Tan, Tieniu},
  journal={International Journal of Computer Vision},
  volume={133},
  number={1},
  pages={31--64},
  year={2025},
  publisher={Springer}
}

@book{pearl2009causality,
  title={Causality},
  author={Pearl, Judea},
  year={2009},
  publisher={Cambridge university press}
}

@book{peters2017elements,
  title={Elements of causal inference: foundations and learning algorithms},
  author={Peters, Jonas and Janzing, Dominik and Sch{\"o}lkopf, Bernhard},
  year={2017},
  publisher={The MIT press}
}

@article{sachs2005,
author = {Karen Sachs  and Omar Perez  and Dana Pe'er  and Douglas A. Lauffenburger  and Garry P. Nolan },
title = {Causal Protein-Signaling Networks Derived from Multiparameter Single-Cell Data},
journal = {Science},
volume = {308},
number = {5721},
pages = {523-529},
year = {2005},
doi = {10.1126/science.1105809},
URL = {https://www.science.org/doi/abs/10.1126/science.1105809},
eprint = {https://www.science.org/doi/pdf/10.1126/science.1105809},
}

@inproceedings{085yu2019dag,
  title={DAG-GNN: DAG structure learning with graph neural networks},
  author={Yu, Yue and Chen, Jie and Gao, Tian and Yu, Mo},
  booktitle={International conference on machine learning},
  pages={7154--7163},
  year={2019},
  organization={PMLR}
}

@inproceedings{
Lachapelle2020Gradient-Based,
title={Gradient-Based Neural DAG Learning},
author={Sébastien Lachapelle and Philippe Brouillard and Tristan Deleu and Simon Lacoste-Julien},
booktitle={International Conference on Learning Representations},
year={2020},
url={https://openreview.net/forum?id=rklbKA4YDS}
}

@article{barabasi2009scale,
  title={Scale-free networks: a decade and beyond},
  author={Barab{\'a}si, Albert-L{\'a}szl{\'o}},
  journal={science},
  volume={325},
  number={5939},
  pages={412--413},
  year={2009},
  publisher={American Association for the Advancement of Science}
}

@article{gilbert1959random,
  title={Random graphs},
  author={Gilbert, Edgar N},
  journal={The Annals of Mathematical Statistics},
  volume={30},
  number={4},
  pages={1141--1144},
  year={1959},
  publisher={JSTOR}
}

@inproceedings{rahimi2007random,
author = {Rahimi, Ali and Recht, Benjamin},
title = {Random features for large-scale kernel machines},
year = {2007},
booktitle = {Proceedings of the 21st International Conference on Neural Information Processing Systems},
pages = {1177–1184},
numpages = {8},
location = {Vancouver, British Columbia, Canada},
}

@article{van2006syntren,
  title={SynTReN: a generator of synthetic gene expression data for design and analysis of structure learning algorithms},
  author={Van den Bulcke, Tim and Van Leemput, Koenraad and Naudts, Bart and van Remortel, Piet and Ma, Hongwu and Verschoren, Alain and De Moor, Bart and Marchal, Kathleen},
  journal={BMC bioinformatics},
  volume={7},
  number={1},
  pages={43},
  year={2006},
  publisher={Springer}
}

@article{peters2014causal,
  title={Causal discovery with continuous additive noise models},
  author={Peters, Jonas and Mooij, Joris M and Janzing, Dominik and Sch{\"o}lkopf, Bernhard},
  journal={The Journal of Machine Learning Research},
  volume={15},
  number={1},
  pages={2009--2053},
  year={2014},
  publisher={JMLR. org}
}

@article{montagna2024demystifying,
  title={Demystifying amortized causal discovery with transformers},
  author={Montagna, Francesco and Cairney-Leeming, Max and Sridhar, Dhanya and Locatello, Francesco},
  journal={arXiv preprint arXiv:2405.16924},
  year={2024}
}

@inproceedings{zhang2009pnl,
author = {Zhang, Kun and Hyv\"{a}rinen, Aapo},
title = {On the identifiability of the post-nonlinear causal model},
year = {2009},
isbn = {9780974903958},
publisher = {AUAI Press},
address = {Arlington, Virginia, USA},
pages = {647–655},
numpages = {9},
location = {Montreal, Quebec, Canada},
series = {UAI '09}
}

@inproceedings{montagna2023causal,
  title={Causal discovery with score matching on additive models with arbitrary noise},
  author={Montagna, Francesco and Noceti, Nicoletta and Rosasco, Lorenzo and Zhang, Kun and Locatello, Francesco},
  booktitle={Conference on Causal Learning and Reasoning},
  pages={726--751},
  year={2023},
  organization={PMLR}
}

@inproceedings{rolland2022score,
  title={Score matching enables causal discovery of nonlinear additive noise models},
  author={Rolland, Paul and Cevher, Volkan and Kleindessner, Matth{\"a}us and Russell, Chris and Janzing, Dominik and Sch{\"o}lkopf, Bernhard and Locatello, Francesco},
  booktitle={International Conference on Machine Learning},
  pages={18741--18753},
  year={2022},
  organization={PMLR}
}

@article{yuksekgonul2026learningdiscovertesttime,
  title={Learning to discover at test time},
  author={Yuksekgonul, Mert and Koceja, Daniel and Li, Xinhao and Bianchi, Federico and McCaleb, Jed and Wang, Xiaolong and Kautz, Jan and Choi, Yejin and Zou, James and Guestrin, Carlos and others},
  journal={arXiv preprint arXiv:2601.16175},
  year={2026}
}

@article{chen2026ticl,
  title={Test-Time Learning of Causal Structure from Interventional Data},
  author={Chen, Wei and Ding, Rui and Huang, Bojun and Zhang, Yang and Fu, Qiang and Liang, Yuxuan and Shi, Han and Zhang, Dongmei},
  journal={arXiv preprint arXiv:2602.19131},
  year={2026}
}

@ARTICLE{Bradley2021,
    
AUTHOR={Butcher, Bradley  and Huang, Vincent S.  and Robinson, Christopher  and Reffin, Jeremy  and Sgaier, Sema K.  and Charles, Grace  and Quadrianto, Novi },
           
TITLE={Causal Datasheet for Datasets: An Evaluation Guide for Real-World Data Analysis and Data Collection Design Using Bayesian Networks},
          
JOURNAL={Frontiers in Artificial Intelligence},
          
VOLUME={Volume 4 - 2021},
  
YEAR={2021},
  
URL={https://www.frontiersin.org/journals/artificial-intelligence/articles/10.3389/frai.2021.612551},
  
DOI={10.3389/frai.2021.612551},
  
ISSN={2624-8212},
}


\appendix

\section{Detailed configuration}
\label{config detail}

\subsection{Setting of static training data}
For all static SCL training setups evaluated(including i.i.d., Graph/Noise/Mechanism shift), we use a total of $K=2,000$ synthetic training instances. Each instance contains $n=200$ i.i.d. observations. The specific training settings for different test instances are as follows:

\begin{table}[htbp]
\caption{{Graph/Noise/Mechanism shift training data setting}}
\label{categorical training}
\resizebox{\textwidth}{!}{
\small
\begin{tabular}{lllllll}
\toprule
& RFF\_G\_ER & RFF\_G\_SF & Linear\_U\_ER & Linear\_U\_SF & Chebyshev\_G\_ER & Chebyshev\_G\_SF \\
\midrule

Graph shift   & RFF\_G\_SF & RFF\_G\_ER & Linear\_U\_SF & Linear\_U\_ER & Chebyshev\_G\_SF & Chebyshev\_G\_ER \\

Noise shift  & RFF\_U\_ER & RFF\_U\_SF & Linear\_L\_ER & Linear\_L\_SF & Chebyshev\_U\_ER & Chebyshev\_U\_SF   \\
Mechanism shift & Chebyshev\_G\_ER & Chebyshev\_G\_SF & RFF\_U\_ER & RFF\_U\_SF & RFF\_G\_ER & RFF\_G\_SF   \\
\bottomrule
\end{tabular}
}
\end{table}

In the Component-mixed setting, these 2,000 instances are uniformly distributed across the 6 mechanism\_noise\_graph combinations, resulting in approximately 330 instances per specific combination. The training data is a mixture of RFF\_U\_ER, RFF\_U\_SF, Linear\_G\_ER, Linear\_G\_SF, Chebyshev\_U\_ER, and Chebyshev\_U\_SF. This makes the model see all components, mechanism (RFF, Linear, Chebyshev), graph (ER, SF), noise (G, U), but not see the specific combination in the test instance, such as RFF\_G\_ER.

\subsection{Hyperparameter settings}

Our score function, $\text{score}(G) = \text{AD}(G, D_{\text{test}}) - \|\mathbf A_G\|_0$, is conceptually aligned with the Akaike Information Criterion (AIC), a well-established model selection criterion formulated as $\text{AIC} = 2k - 2\ln(L)$, where $k$ is the number of parameters and $L$ is the likelihood. In our formulation, $\|\mathbf A_G\|_0$ is analogous to the number of parameters $k$, and the negative log-likelihood $-\log p(X \mid f)$ is analogous to $-\ln(L)$. 

We perform $N_{steps}=2000$ stochastic refinement steps per test instance. Our current $K$ is fixed at 200, and we have achieved good results on all our current datasets. It is foreseeable that performance will further improve as K increases, but this depends on our balance between performance and time consumption. 

\section{Consistency on other model backbones}
\label{other backbone}

To further validate the generality of the TTT-SCL framework and the observed o.o.d generalization challenges across different model architectures, we conduct experiments using the Pairwise Attention from \cite{zhang2025learning} (\textbf{SiCL}) as an alternative model backbone. Unlike the AVICI transformer used in the main experiments, which predicts a full directed adjacency matrix (DAG), SiCL incorporates pairwise attention mechanisms and is trained to predict the undirected skeleton and v-structures of the causal graph. This setup allows us to investigate whether the identified o.o.d failure patterns persist when using a fundamentally different architecture (with pairwise attention) and a different learning target (skeleton and v-structures instead of a full DAG), thereby testing the robustness of our conclusions.

\subsection{Experimental Setup}
Backbone Model is SiCL (Pairwise Attention Network) \cite{zhang2025learning}. Learning Target is Undirected graph skeleton. The training strategy for the static baseline models (i.i.d. and SiCL(mixed)) follows the same data generation procedures described in Section 5.1.1 of the main text, but the ground-truth labels are converted to the appropriate representation for SiCL (skeleton labels). Evaluation Metric is AUROC for edge presence in the predicted skeleton. OOD Settings is identical to those defined for Table 1 in the main text: \textit{i.i.d.}, \textit{Graph shift}, \textit{Noise shift}, \textit{Mechanism shift}. The \textit{AVICI(mixed)} is replaced with \textit{SiCL(mixed)}, respectively.

\subsection{Results and Analysis}
Table~\ref{tab:skeleton_ood_sicl} presents the AUROC for skeleton discovery under different distribution shifts.Consistent with the findings in Fig ~\ref{fig:ood} using the AVICI backbone, the SiCL backbone—which employs a fundamentally different pairwise attention architecture and learns undirected skeletons rather than full DAGs—exhibits the same pattern of out-of-distribution generalization failure. Under i.i.d. conditions, SiCL achieves perfect or near-perfect performance. However, significant performance degradation occurs across all types of distribution shifts, with mechanism shifts proving particularly damaging (e.g., dropping to 66.5 on RFF\_G\_SF and 58.4 on Chebyshev\_G\_SF). Critically, the SiCL(mixed) variant, while trained on data containing all individual distributional components (graph types, mechanisms, and noise distributions), still fails to generalize to novel combinations of these factors. This demonstrates that SCL models struggle with compositional generalization—they memorize specific configuration patterns rather than learning modular causal representations. These results demonstrate that the OOD generalization challenge is not specific to a particular model architecture or output representation, but represents a fundamental limitation of the static pre-training paradigm in supervised causal learning. The consistent failure patterns across both transformer-based (AVICI) and pairwise-attention-based (SiCL) models strongly validate the need for test-time adaptation frameworks like TTT-SCL.

\begin{table}[h!]
\centering
\caption{OOD generalization performance for skeleton using the SiCL (Pairwise Attention) backbone.}
\label{tab:skeleton_ood_sicl}
\resizebox{\textwidth}{!}{
\setlength{\tabcolsep}{4pt}
\begin{tabular}{l c c c c c c} 
\toprule
 & RFF\_G\_ER & RFF\_G\_SF & Linear\_U\_ER & Linear\_U\_SF & Chebyshev\_G\_ER & Chebyshev\_G\_SF \\
\midrule
iid & 82.1(6.7) & 100.0(0.0) & 81.4(6.9) & 100.0(0.0) & 94.3(2.8) & 100.0(0.0) \\
Graph shift & 66.4(9.0) & 85.4(4.1) & 65.8(6.9) & 94.0(2.5) & 73.0(5.7) &  92.9(4.3) \\
Noise shift & 60.0(8.9) & 91.7(3.8) & 65.3(7.4) & 84.0(7.7) & 88.6(5.3) & 89.3(5.4)\\
Mechanism shift & 62.1(7.4) & 66.5(6.3) & 59.4(4.7) & 83.8(4.8) & 76.1(8.9) & 58.4(9.7)\\
SiCL(mixed) & 64.4(8.0) & 74.4(10.7) & 66.7(7.3) & 82.7(8.2) & 85.6(3.7) & 91.2(4.1)\\
\bottomrule
\end{tabular}
}
\end{table}

\section{Structure-Induced Mechanism (SIM)}
\label{app:sim}

\paragraph{Intuition}
Given an observed test dataset \(D_{test}=\{x^{(t)}\}_{t=1}^{n}\) and a candidate DAG \(G_k\), SIM (structure-induced mechanism)
produces a synthetic dataset \(D_k\) whose distribution is induced by the pair \((G_k,\Theta_k)\),
where \(\Theta_k\) denotes the parameterized mechanisms (and noise laws) fitted from \(D_{\mathrm{test}}\).

\paragraph{Formal description.}
Assume variables \(X_1,\dots,X_d\). For each node \(i\) with parent set \(\mathrm{Pa}_{G_k}(X_i)\),
we estimate a mechanism \(f_{k,i} \in \mathcal{F}\) and a noise distribution \(P_{\epsilon_{k,i}}\) by solving
\[
\hat f_{k,i} \;=\; \arg\min_{f\in\mathcal{F}} \;\sum_{t=1}^n \ell\big(x_i^{(t)},\, f\big( x_{\mathrm{Pa}_{G_k}(X_i)}^{(t)}\big)\big)
\]
where \(\ell(\cdot,\cdot)\) is a suitable loss (e.g. squared error for continuous targets) and \(x_{\mathrm{Pa}}^{(t)}\) denotes observed parent-values
in sample \(t\). Residuals are computed as
\[
\hat\epsilon_{k,i}^{(t)} = x_i^{(t)} - \hat f_{k,i}\big( x_{\mathrm{Pa}_{G_k}(X_i)}^{(t)}\big).
\]
The fitted mechanism set is denoted \(\Theta_k = \{\hat f_{k,i}, \,\hat P_{\epsilon_{k,i}}\}_{i=1}^d\),
where \(\hat P_{\epsilon_{k,i}}\) is obtained either parametrically (e.g. Gaussian fit) or non-parametrically
(empirical residual distribution / bootstrap). Once \(\Theta_k\) is obtained, generate \(n\) synthetic samples
\(\tilde x^{(1)},\dots,\tilde x^{(n)}\) by ancestral (topological) sampling:
for \(t=1,\dots,n\) and nodes in topological order set
\[
\tilde X_i^{(t)} = \hat f_{k,i}\big( \tilde X_{\mathrm{Pa}_{G_k}(X_i)}^{(t)} \big) + \tilde\epsilon_{k,i}^{(t)},
\]
where \(\tilde\epsilon_{k,i}^{(t)}\sim \hat P_{\epsilon_{k,i}}\). The resulting synthetic dataset is \(D_k=\{\tilde x^{(t)}\}_{t=1}^n\).

\section{Implementation of AD}
\label{ADs}
In the main text, we propose the Alignment of Distribution (AD) metric as a core measure of causal similarity between the generated training data $D_{train}$ and the test instance $D_{test}$. While the likelihood-based implementation was used in our primary experiments, we provide alternative formulations here to accommodate different data distributions and modeling assumptions.

\subsection{$R^2$-based AD}
For continuous variables under additive noise models, the coefficient of determination ($R^2$) provides an intuitive measure of goodness-of-fit for each causal mechanism:
\[
AD_{R^2}(G_{train}, D_{test}) = \frac{1}{d} \sum_{i=1}^{d} \left[ \frac{1}{K} \sum_{k=1}^{K} R^2\left( f_i^k(\mathbf{Pa}^k(X_i)), X_i \right) \right]
\]
This value approaches 1 when the fitted mechanisms explain the variance in $D_{test}$ well, indicating strong alignment.

\subsection{Normalized Wasserstein Distance-based AD}
For multi-modal or heavy-tailed distributions, the Wasserstein distance offers a robust metric for comparing empirical distributions. We define a \textit{Normalized Wasserstein Distance (NWD)} based AD metric as follows:

For a given variable $X_i$ and a candidate graph $G^k$ with its fitted mechanism $f_i^k$, we compute:
\[
\text{NWD}(f_i^k, G^k, D^{{test}}) := 1 - \frac{W_1\left( \{x_i\},\ \{f_i^k (\mathbf{Pa}^k(X_i))\} \right)}{\text{max}(\mathcal{U}) - \text{min}(\mathcal{U})}
\]
where:
\begin{itemize}
    \item $\{x_i\}$ are the observed values of $X_i$ in $D_{{test}}$.
    \item $\{f_i^k (\mathbf{Pa}^k(X_i))\}$ are the values generated by applying the fitted mechanism $f_i^k$ to the parent values in $D_{{test}}$.
    \item $W_1$ is the 1-Wasserstein distance (Earth Mover's Distance). For two equally sized, sorted collections of values $\{a^{(j)}\}$ and $\{b^{(j)}\}$, it is defined as:
    \[
    W_1(\{a\}, \{b\}) = \frac{1}{n} \sum_{j=1}^n |a^{(j)} - b^{(j)}|
    \]
    \item $\mathcal{U} = \{x_i\} \cup \{f_i^k (\mathbf{Pa}^k(X_i))\}$ is the union of the observed and generated values for $X_i$.
    \item The denominator, $\text{max}(\mathcal{U}) - \text{min}(\mathcal{U})$, is the range of the combined set, used for normalization.
\end{itemize}
The resulting NWD value lies between 0 and 1, where 1 indicates a perfect match between the generated and observed distributions for that variable. The overall AD metric is then the average NWD across all variables and generated graphs:
\[
AD_{\text{NWD}}(G_{train}, D_{test}) = \frac{1}{K} \sum_{k=1}^{K} \left[ \frac{1}{d} \sum_{i=1}^{d} \text{NWD}(f_i^k, G^k, D^{{test}}) \right]
\]

\subsection{Selection Guidance}
The \textbf{likelihood-based} AD is most natural for probabilistic models and was used in our main experiments. The \textbf{$R^2$-based} AD is suitable for continuous variables under additive noise assumptions, often leading to computationally efficient and intuitive scores. The \textbf{NWD-based} AD is recommended for complex, non-Gaussian, or heavy-tailed distributions where likelihood or $R^2$ might be less informative or robust.
The TTT-SCL framework is agnostic to the specific choice of AD metric, allowing users to select the most appropriate one for their domain.

\section{Workflow of Training Sets Generation}
\label{Workflow of training sets generation}

To further illustrate the generation of training sets, we have drawn a workflow to visualize the process, as shown in Fig. ~\ref{training sets generation}.

\begin{figure}[htb]
\centering
\includegraphics[width=\linewidth]{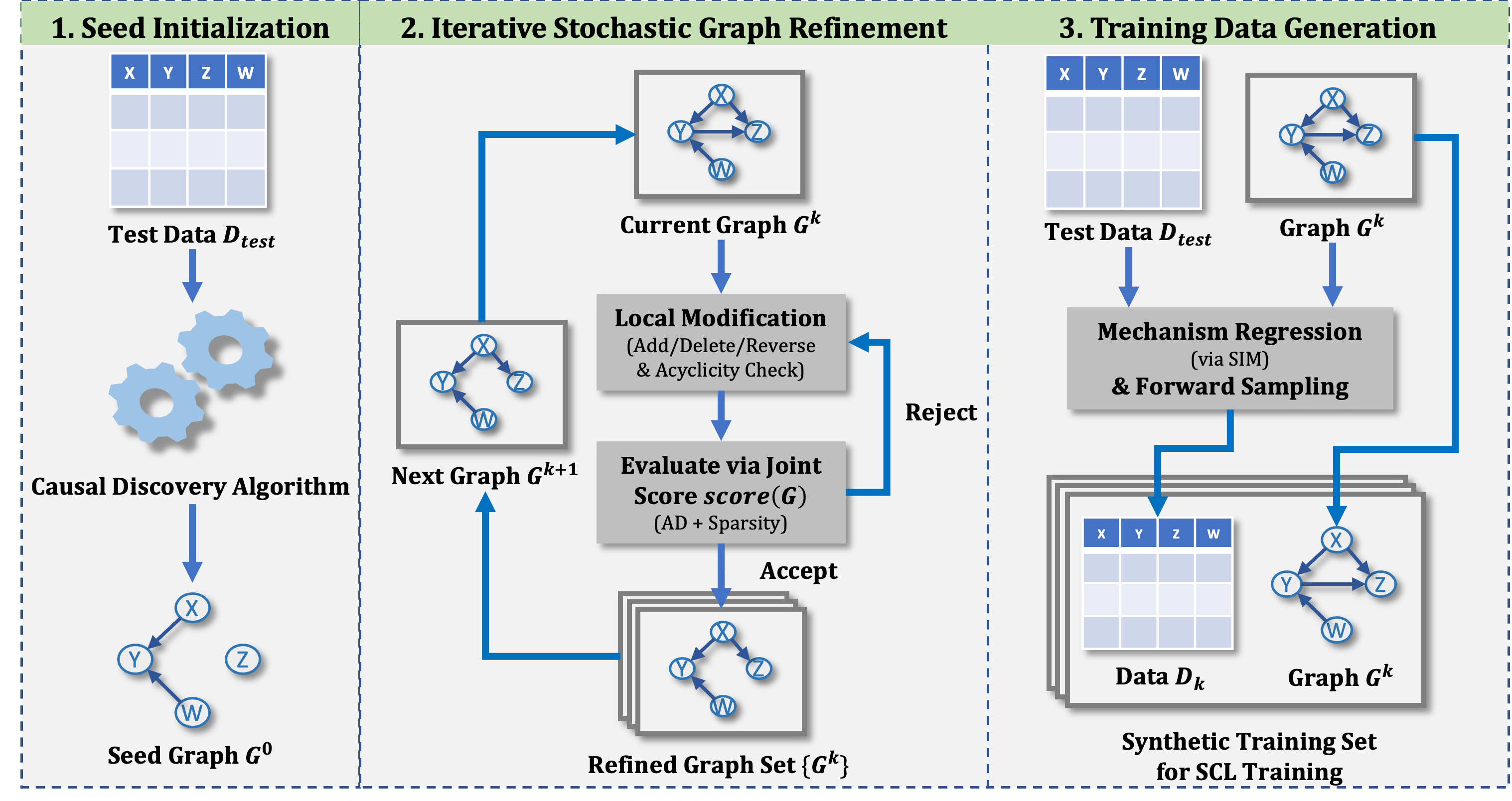}
    \caption{Workflow of training sets generation}
    \label{training sets generation}
\end{figure}

\section{Performance in Other Metrics}
\label{other metrics}

In the main text, we primarily reported the AUROC for edge prediction to succinctly demonstrate the impact of training data quality on model performance. For a more comprehensive evaluation, we provide results on additional standard causal discovery metrics in this appendix:
\begin{itemize}[left=0pt]

    \item \textbf{Accuracy (ACC)}: The proportion of correctly predicted edge presence/absence across all possible edges. Higher is better. This metric can be viewed as a normalized version of the Structural Hamming Distance (SHD), where instead of counting the number of incorrect edges, it measures the proportion of correct edge predictions relative to the total possible edges.
    \item \textbf{F1-Score}: The harmonic mean of precision and recall for edge prediction. Higher is better.
    \item \textbf{Area Under the Precision-Recall Curve (AUPRC)}: Particularly informative under class imbalance (sparse graphs). Higher is better.
\end{itemize}

\begin{table}[h!]
\centering
\caption{Comprehensive evaluation across multiple datasets and metrics. Mean (standard deviation) over multiple runs are reported for synthetic data. Best results are in \textbf{bold}.}
\label{tab:metrics_comprehensive}
\resizebox{\textwidth}{!}{
\begin{tabular}{l l l l l l l l l l l l l l l l}
\toprule
\multirow{3}{*}{Method} & \multicolumn{3}{c}{RFF\_G} & \multicolumn{3}{c}{Linear\_U} & \multicolumn{3}{c}{Chebyshev\_G} & \multicolumn{3}{c}{Sachs} & \multicolumn{3}{c}{Syntren} \\
\cmidrule(lr){2-4} \cmidrule(lr){5-7} \cmidrule(lr){8-10} \cmidrule(lr){11-13} \cmidrule(lr){14-16}
 & ACC$\uparrow$ & F1$\uparrow$ & AUPRC$\uparrow$ & ACC$\uparrow$ & F1$\uparrow$ & AUPRC$\uparrow$ & ACC$\uparrow$ & F1$\uparrow$ & AUPRC$\uparrow$ & ACC$\uparrow$ & F1$\uparrow$ & AUPRC$\uparrow$ & ACC$\uparrow$ & F1$\uparrow$ & AUPRC$\uparrow$ \\
\midrule
PC & 75.6(4.2) & 39.5(9.7) & 37.5(6.3) & 74.0(4.4) & 40.4(8.4) & 37.2(6.2) & 73.7(5.1) & 37.4(12.6) & 36.8(7.8) & 84.2 & 45.7 & 30.1 & 84.75 & 16.43 & 6.89 \\
GES & 76.7(7.7) & 49.8(16.3) & 43.5(12.3) & 71.9(9.8) & 55.4(12.9) & 45.3(9.3) & 72.1(5.1) & 38.5(10.5) & 35.7(6.6) & 82.6 & 36.3 & 24.2 & 65.50 & 1.42 & 4.79 \\
NOTEARS & 86.6(3.3) & 73.2(6.2) & 64.1(7.4) & 89.1(2.9) & 76.6(7.3) & 69.7(8.6) & 72.3(2.7) & 14.5(8.7) & 29.8(3.3) & 82.6 & 36.3 & 24.2 & \textbf{94.75} & 0.00 & 5.00 \\
AVICI(scm-v0) & \textbf{93.1(1.6)} & \textbf{87.3(3.4)} & \textbf{94.9(3.1)} & 73.9(7.1) & 41.8(18.4) & 52.8(17.0) & 80.6(5.4) & 58.4(14.5) & 69.3(14.2) & 83.4 & 23.0 & 31.6 & 93.00 & 22.22 & 25.53 \\
TTT-SCL (random) & 83.2(5.3) & 72.8(9.4) & 68.8(10.8) & 75.9(6.1) & 59.8(11.8) & 56.2(11.5) & 75.5(7.0) & 56.6(10.8) & 60.0(10.0) & 68.5 & 24.0 & 24.5 & 72.50 & 16.66 & \textbf{53.91} \\
TTT-SCL (Notears) & 86.8(3.5) & 78.4(6.1) & 76.0(8.6) & \textbf{78.7(3.9)} & \textbf{65.4(8.0)} & \textbf{65.0(9.9)} & \textbf{77.1(6.7)} & \textbf{61.9(10.2)} & \textbf{66.0(16.3)} & \textbf{85.9} & \textbf{56.4} & \textbf{53.6} & 90.50 & \textbf{32.14} & 51.85 \\
\bottomrule
\end{tabular}%
}
\end{table}

Table~\ref{tab:metrics_comprehensive} presents the performance of all compared methods across three distinct synthetic data settings (RFF\_G, Linear\_U, and Chebyshev\_G) and the real-world Sachs dataset. TTT-SCL (Notears) achieves highly competitive performance across all datasets and evaluation metrics (ACC, F1, AUPRC), demonstrating its robustness to distribution shifts. It consistently outperforms traditional methods (PC, GES, NOTEARS) and the strong pre-trained SCL baseline AVICI(scm-v0) on most settings, particularly on the challenging Chebyshev\_G and real-world Sachs dataset. While AVICI(scm-v0) excels in the RFF\_G setting it was trained on, its performance degrades significantly under mechanism shifts (Linear\_U) and on real data, highlighting the limitation of static pre-training. The superior performance of TTT-SCL across multiple metrics confirms that its test-time training strategy generates high-quality, causally-aligned training data, leading to more accurate and reliable causal discovery.

\section{Additional Experiments on Benchmark Causal Graphs}
\label{app:bnlearn_experiments}

To further validate TTT-SCL's performance on real-world causal structures, we conducted additional experiments on well-established benchmark causal graphs from the bnlearn repository. The scarcity of real-world causal datasets with ground truth is a fundamental challenge in causal discovery research. While most works primarily rely on synthetic data and a limited number of real datasets (e.g., Sachs), benchmark causal graphs from bnlearn provide valuable testbeds as they represent causal structures derived from real-world domains and expert knowledge.

We selected four representative graphs from bnlearn:
\begin{itemize}[left=0pt]
    \item \textbf{Asia:} A classic medical diagnostic network modeling the relationships between visiting Asia, smoking, tuberculosis, lung cancer, bronchitis, and various test results. This graph represents a well-known benchmark in causal inference with clear medical relevance.
    
    \item \textbf{Cancer:} A compact but meaningful graph modeling causal relationships in cancer epidemiology, including pollution, smoking, and genetic factors. Its small size belies its representativeness of real-world medical causal reasoning.
    
    \item \textbf{Earthquake:} Models causal relationships between burglary, earthquake, alarm triggers, and neighbor responses. This graph exemplifies causal reasoning in security and monitoring systems.
    
    \item \textbf{Survey:} Represents causal relationships in social science research, including age, sex, education, occupation, and transportation preferences. This graph demonstrates causal structures in sociological studies.
\end{itemize}

These benchmark graphs are representative because they: (1) capture diverse real-world domains (medical, social, security), (2) are widely recognized and validated in the causal inference literature, and (3) reflect expert-curated causal knowledge rather than purely synthetic constructions.

We parameterized these graphs using Chebyshev polynomial mechanisms to generate pseudo-real datasets, maintaining the authentic causal structures while incorporating realistic nonlinear relationships. Table~\ref{tab:bnlearn_results} shows that TTT-SCL consistently achieves state-of-the-art performance across all benchmark graphs, demonstrating its robustness to diverse real-world causal structures.

\begin{table}[h]
  \centering
  \caption{{Performance comparison (AUROC) on benchmark causal graphs from bnlearn repository.}}
  \label{tab:bnlearn_results}
  \begin{tabular}{l l l l l}
    \toprule
    Method & Asia & Cancer & Earthquake & Survey \\
    \midrule
    PC & 74.1 & 70.2 & 75.5 & 90.0 \\
    GES & 46.4 & 85.1 & 80.3 & 88.3 \\
    NOTEARS & 68.7 & 87.5 & 60.1 & 64.9 \\
    AVICI (scm-v0) & 83.3 & 86.9 & 94.0 & 89.4 \\
    TTT-SCL (random) & 86.8 & 84.5 & 84.5 & 92.7 \\
    TTT-SCL (NOTEARS) & \textbf{91.0} & \textbf{91.6} & \textbf{98.8} & \textbf{95.5} \\
    \bottomrule
  \end{tabular}
\end{table}

The superior performance of TTT-SCL across these diverse benchmark graphs further validates its effectiveness in handling real-world causal structures.

\section{Sensitivity to Seed Graph Initialization}
\label{app:seed}

As shown in Table~\ref{tab:performance_TTT-SCL}, TTT-SCL (Notears) consistently outperforms TTT-SCL (random), raising the question of sensitivity to initialization. We argue that this gap reflects the finite search budget rather than a fundamental limitation. To verify this, we introduce a variant TTT-SCL (random-long) that uses a random seed but extends the stochastic refinement steps to $N_{\text{steps}} = 5000$ (vs.\ the default $2000$). 

\begin{table}[htbp]
    \centering
    \caption{Effect of seed quality and search budget on AUROC.}
    \label{tab:seed}
    \begin{tabular}{l c c}
        \toprule
        & RFF\_G & Sachs \\
        \midrule
        TTT-SCL (Notears)     & 91.8 & 78.9 \\
        TTT-SCL (random)      & 88.4 & 58.6 \\
        TTT-SCL (random-long) & 89.1 & 71.9 \\
        \bottomrule
    \end{tabular}
\end{table}

The results in Table ~\ref{tab:seed} clearly show that with a larger computational budget, the performance gap narrows considerably. On the Sachs dataset, TTT-SCL (random-long) reaches 71.9 AUROC, a substantial improvement over 58.6, and approaches the 78.9 achieved with a Notears seed. This confirms that the stochastic refinement process is effective on its own, and a poorer seed can be compensated by additional search steps. Therefore, using a strong initializer like Notears is a practical recommendation for efficiency, not a strict requirement.

\section{Ablation study}
\label{Ablation study}

The main text established the necessity of the sparsity constraint in the TTT-SCL optimization objective to prevent degenerate, overly dense solutions. This appendix provides further empirical evidence to dissect the roles of the AD metric and the sparsity constraint. 

\subsection{The role of AD and sparsity}

To further demonstrate the effectiveness of AD and the necessity of sparsity, the AD, sparsity, score of the training data obtained by different methods under different test data, as well as the AUROC on the test data were recorded in Table \ref{tab:data_quality}. The combined optimization of AD and sparsity is critical for generating high-quality training data. Without sparsity constraints (TTT-SCL(Notears-s)), high AD values alone lead to overly dense graphs that overfit the test distribution, violating causal minimality and resulting in lower AUROC. In contrast, jointly optimizing AD and sparsity (TTT-SCL(Notears)) yields training data that is both distributionally aligned and structurally sparse, closely matching the true causal graph. The resulting composite score strongly correlates with final model AUROC, confirming that both components are essential for robust generalization under distribution shifts, especially mechanism shifts. 

\begin{table}[htbp]
\centering
\caption{{AD and Sparsity characterize the quality of the training data.}}
\label{tab:data_quality}
\resizebox{0.8\textwidth}{!}{
\begin{tabular}{l l c c c}
\toprule
Metric & Methods & RFF\_G & Linear\_U & Chebyshev\_G \\
\midrule
\multirow{3}{*}{AD}       & TTT-SCL(random)    & -370.0 & -258.5 & -303.5 \\
                          & TTT-SCL(Notears-s) & -357.5 & -217.5 & -298.0 \\
                          & TTT-SCL(Notears)   & -363.0 & -220.5 & -308.0 \\
\addlinespace[2pt]
\midrule
\multirow{3}{*}{Sparsity} & TTT-SCL(random)    & 33.15  & 34.00  & 29.49  \\
                          & TTT-SCL(Notears-s) & 38.80  & 38.75  & 38.85  \\
                          & TTT-SCL(Notears)   & 31.95  & 35.65  & 27.25  \\
\addlinespace[2pt]
\midrule
\multirow{3}{*}{Score}    & TTT-SCL(random)    & -403.3 & -293.0 & -333.3 \\
                          & TTT-SCL(Notears-s) & -397.0 & -256.8 & -337.8 \\
                          & TTT-SCL(Notears)   & -395.0 & -256.5 & -335.5 \\
\addlinespace[2pt]
\midrule
\multirow{3}{*}{AUROC}    & TTT-SCL(random)    & 0.884  & 0.823  & 0.796  \\
                          & TTT-SCL(Notears-s) & 0.868  & 0.843  & 0.697  \\
                          & TTT-SCL(Notears)   & 0.918  & 0.863  & 0.830  \\
\bottomrule
\multicolumn{5}{l}{\footnotesize Note: AD/Score/AUROC (Higher is better), Sparsity (Low is better)} \\
\end{tabular}
}
\end{table}

\subsection{Control AD, change sparsity}
To control sparsity independent of AD, we design a controlled experiment based on the ground-truth test graph $G_{test}$. For a given $G_{test}$ and its observational data $D_{test}$, we generate alternative candidate training graphs $G_{train}$ by \textbf{gradually adding extra edges} to $G_{test}$ (while ensuring the resulting graph remains a DAG). This creates a series of graphs that are supergraphs of the true graph.

\begin{itemize}[left=0pt]
    \item \textbf{Setting 1 (Sparse)}: Add a small number of extra edges ($|E_{add}| = m_1$).
    \item \textbf{Setting 2 (Medium)}: Add a medium number of extra edges ($|E_{add}| = m_2$, $m_2 > m_1$).
    \item \textbf{Setting 3 (Dense)}: Add a large number of extra edges ($|E_{add}| = m_3$, $m_3 > m_2$).
\end{itemize}

For each generated supergraph $G_{train}$ in these settings, we then:
1.  Parameter Fitting: Regress the mechanisms $f_i$ and noise distributions from $D_{test}$ using $G_{train}$ (via SIM).
2.  Forward Sampling: Generate synthetic training data $D_{train}$ from the fitted SCM $(G_{train}, f, \epsilon)$.
3.  Calculate Metrics: Compute the AD score between $D_{train}$ and $D_{test}$, and the sparsity of $G_{train}$.
4.  Train \& Evaluate: For each $(G_{train}, D_{train})$ pair, train an SCL model (AVICI backbone) and evaluate its AUROC on recovering the \textit{true} $G_{test}$ from $D_{test}$.

This procedure is repeated for $K$ graphs per setting. The key insight is that by construction, all generated $G_{train}$ graphs are capable of representing the data distribution $D_{test}$. Therefore, we expect them to achieve similar, high AD scores. However, only the sparsest graph ($G_{test}$ itself) represents the true causal structure.

Table~\ref{tab:control_ad_sparsity} shows the results for the \textbf{RFF\_ER\_G} dataset, which are representative of the overall trend.

\begin{table}[htbp]
  \centering
  \caption{Control AD, change sparsity} %
  \label{tab:control_ad_sparsity}
  \begin{tabular}{l l l l l}
    \toprule
    RFF\_ER\_G & setting & AD & sparsity & AUROC \\
    \midrule
    \multirow{3}{*}{Control AD, change sparsity} 
    & 1 & -375 & 25.91 & 1.0(0) \\
    & 2 & -368 (+1.8\%) & 32.32(+24.7\%) & 0.972(0.017) \\
    & 3 & -362(+3.4\%) & 36.59(+41.2\%) & 0.908(0.023) \\
    \bottomrule
  \end{tabular}
\end{table}

The results clearly demonstrate the critical, independent role of the sparsity constraint. All supergraphs achieve a high and similar AD score (variation $< 4\%$), confirming that many different graphs can explain the observed data distribution nearly equally well. This illustrates the identifiability crisis without further constraints. As expected, adding more edges increases the sparsity metric (number of edges). Crucially, the downstream performance (AUROC) of the SCL model \textbf{degrades significantly as the graphs become denser}, even though the AD score remains high. The model trained on the true graph (Setting 1, perfect sparsity) achieves perfect AUROC. Performance drops to 0.972 for medium density and further to 0.908 for high density.
    
\section{Scalability analysis and limitation}
\label{app:runtime}

The computational complexity of TTT-SCL is dominated by the Stochastic Graph Refinement step. The search space for Directed Acyclic Graphs (DAGs) with $d$ variables is super-exponential, rendering exhaustive search intractable. Our stochastic search conducts a guided walk through this space, and its complexity is determined by the number of steps $N_{\text{steps}}$ and the cost of evaluating the Alignment of Distribution (AD) score for each candidate graph. 

The evaluation for a single candidate graph $G$ involves:
\begin{itemize}[left=0pt]
    \item \textbf{Mechanism Fitting}: For each node \(X_i\), we fit a causal mechanism \(f_i\) (using a Generalized Additive Model) based on its parent set \(\mathbf{Pa}_G(X_i)\) from the test data \(D_{\text{test}}\) with $n$ samples. Let \(k_i = |\mathbf{Pa}_G(X_i)|\) be the in-degree of \(X_i\). The cost of fitting for one node is typically \(O(n \cdot k_i \cdot l)\), where $l$ is the number of GAM iterations. Due to the sparsity constraint \(\text{Sparsity}(G)\) in our joint score function (Eq.~5), which actively penalizes dense graphs, the in-degrees $k_i$ encountered during the search are small. Letting $H$ represent the small, approximately constant maximum in-degree enforced by this constraint, the cost per node becomes $O(n \cdot H \cdot l)$. Aggregated across all $d$ nodes, the total fitting cost is $O(n \cdot H \cdot d \cdot l)$, which simplifies to $O(n \cdot d)$ since $H$ and $l$ are constants.
    
    \item \textbf{Likelihood Calculation}: After fitting, computing the log-likelihood for all $n$ samples and $d$ variables has a cost of $O(n \cdot d)$.
\end{itemize}

Thus, the per-step AD evaluation cost is $O(n \cdot d)$. The total complexity of the Stochastic Graph Refinement phase is therefore $O(N_{\text{steps}} \cdot n \cdot d)$. The subsequent Training Data Generation step involves fitting mechanisms and forward-sampling for only the final $K$ selected graphs, contributing a minor additive term of $O(K \cdot n \cdot d)$, which is negligible since $N_{\text{steps}} \gg K$ (in our experiments, $N_{\text{steps}}=2000$ and $K=200$).

To empirically validate this theoretical analysis, we present a runtime breakdown in Table~\ref{tab:runtime_breakdown}. The results confirm that Stochastic Graph Refinement is indeed the computational bottleneck, as it involves thousands of AD evaluations. The subsequent Training Data Generation step, which performs SIM fitting on only the final $K$ selected graphs, constitutes a minor fraction of the total time. Model training time is also relatively modest compared to the graph refinement phase. Our experiment used an 80G A100 GPU, and when the number of nodes was 10, it only used about 4G of memory.

At the same time, we would like to emphasize that causal discovery is fundamentally a scientific inference task where high computational cost is often acceptable in exchange for accurate and reliable results. For example, when exploring the interactions between different proteins ~\citep{sachs2005}, or the relationships between different physiological factors in children with heart disease ~\citep{Bradley2021}, we don't care whether it's 1 minute or 1 hour, or even days or weeks; we want to get a more reliable result.

\begin{table}[htbp]
    \centering
    \caption{{Runtime breakdown of TTT-SCL for a test instance with varying number of nodes.}} 
    \label{tab:runtime_breakdown}
    \begin{tabular}{l l l l}
        \toprule
        Component & 10 nodes & 20 nodes & 30 nodes \\
        \midrule
        Stochastic Graph Refinement & 26 min & 61 min & 113 min \\
        Training Data Generation (SIM fitting) & 1.3 min & 3.2 min & 5.6 min \\
        Model Training & 3.3 min & 5.8 min & 8.3 min \\
        \bottomrule
    \end{tabular}
\end{table}

\end{document}